\def\year{2015}
\documentclass[letterpaper]{article}
\usepackage{graphicx}
\usepackage{aaai}
\usepackage{times}
\usepackage{helvet}
\usepackage{courier}

\usepackage{latexsym}
\usepackage{epic}
\usepackage{amsfonts}
\usepackage{amsmath}
\usepackage{amssymb}
\usepackage{amsthm}
\usepackage{xspace}

\usepackage{float}
\floatstyle{boxed}
\restylefloat{figure}

\newtheorem{myexample}{Example}

\newcommand{\myproof}{\noindent {\bf Proof:\ \ }}

\newtheorem{mytheorem}{Theorem}

\newtheorem{remark}{Remark}

\newcommand{\myqed}{\mbox{$\diamond$}}

\newcommand{\myOmit}[1]{}

\begin{document}
%
\title{Welfare of Sequential Allocation Mechanisms for Indivisible Goods}

\author{Haris Aziz \\
Data61 and UNSW,\\
 Sydney 2033, Australia
 \And Thomas Kalinowski\\
 University of Newcastle\\
 Newcastle, Australia
\And Toby Walsh\\
UNSW and Data61,\\
 Sydney 2033, Australia
 \And Lirong Xia\\
RPI\\
NY 12180, USA
}

\maketitle
\begin{abstract}
Sequential allocation is a simple and attractive
mechanism for the allocation of indivisible goods. 
Agents take
turns, according to a policy, to pick items. 
Sequential allocation is guaranteed to return
an allocation which is efficient but may not
have an optimal social welfare. 
We consider therefore the relation between welfare
and efficiency. We study the (computational) questions
of what welfare is possible or necessary
depending on the choice of policy. 
We also consider a novel control problem in which the
chair chooses a policy to improve
social welfare. 
\end{abstract}

\section{Introduction}

Due to economical, environmental and political concerns,
we often want to do more with 
fewer resources and to do so more fairly. 
One way to achieve this is to use 
computing power to improve 
the efficiency and equitability of the allocation. 
One important and challenging case is the
fair division of indivisible goods. This
captures a wide range of problems
including allocating classes to students,
landing slots to airlines, players to teams, and houses to
people. 

A simple but popular mechanism to allocate
indivisible goods is {\em sequential allocation} 
\cite{cakecut}. 
Agents simply take turns to pick items. 
The sequential allocation mechanism leaves open the particular
order used to take turns (the so called ``policy'').
Is it fairest perhaps to have a balanced alternating
policy in which items are allocated in rounds, 
each agent picks one item in each round, but we reverse the order
of the agents after each round. 
Indeed, there are real world settings like
course allocation at the Harvard Business School where
the policy is chosen at random from a space
of balanced alternating policies as a means of ensuring (procedural)
fairness. 

The choice of policy impacts the welfare of the resulting
allocation. 
This raises the question of what social welfare can or must
be achieved. Do we {\em necessarily} achieve a minimum
acceptable welfare whatever policy is
chosen? Is is {\em possible} 
that the welfare is above a required minimum?
What is the {\em maximum} welfare that
can be achieved? 
What is the {\em minimum} welfare that will 
be achieved? These questions
are closely related to an interesting control problem. 
Can a (benevolent) chair choose a policy to improve or
maximize welfare? These questions are also related
to the expected welfare when the
policy is chosen at random. 
For example, the expected welfare 
is between the minimum welfare
that is necessary and the 
maximum welfare that is possible
(strictly so when they are different). 

We are not the first to consider
what allocations are possible or necessary
depending on the choice of policy. 
Aziz, Walsh and Xia \shortcite{awxaaai15}
ask what item or (sub)set of items can or will be
allocated depending on the choice of policy.
By comparison, we consider here not the items allocated
but the welfare achieved. 
We are also not the first to consider the policy
that maximizes welfare. 
For example, Bouveret and Lang \shortcite{indivisible}
ask what is the ``optimal''
policy that maximizes the {\em expected} egalitarian
or utilitarian social welfare.
However, their model supposes that the
ordinal preferences are not known by the chair
and optimality is in expectation under two extreme assumptions:
full independence (all rankings equiprobable) and
full correlation (identical rankings). 
By comparison, we consider here the case
where the we know the {\em exact} utilities and can
therefore maximize the {\em actual} social welfare.

Sequential allocation is an ordinal mechanism
(it merely requires agents to declare an
ordering over items). In many of our settings, however,
we suppose that we know the agents' utilities. 
This may be because
we know the ordinal preferences
but utilities can be easily
computed from these (e.g. Borda or lexicographical utilities). 
In other cases, we might suppose that we have elicited the agents'
general utilities, and we then compute a policy
to maximize welfare which we announce and use
to allocate items using sequential
allocation. Even in this more complex
setting, we retain some of the advantages of
a purely ordinal mechanism. For instance,
it is easy for the agents to verify that
the policy is fair (e.g. it is a balanced alternating
policy), and the part of the mechanism allocating items
according to the declared policy has been applied 
correctly. 

\section{Welfare and efficiency}

We first consider the precise relationship
between social welfare and efficiency.
We suppose that there are $n$ agents being allocated
$m$ items. Agents have additive utilities
over the items. Agents convert these into 
a strict ordinal ranking over items, breaking
any ties in utility in some fixed way. 
The welfare of an agent is simply 
the sum of the utilities of the items allocated to that
agent. The utilitarian welfare is the sum of the welfare of the agents, 
whilst the egalitarian welfare is that of
the worst off agent (or agents). 
The sequential allocation mechanism is parameterized
by the policy, the order in which agents pick items.
For example, with
the policy $123321$,
agent 1 picks first, then agent 2, then agent 3
before we repeat in reverse. 
An allocation is an assignment of items to agents.
One allocation {\em Pareto improves} another iff 
each agent has at least the same utility in the first, and 
there is at least one agent where the utility is greater. 
An allocation is {\em Pareto efficient} iff 
there is no allocation which Pareto improves it. 
For every Pareto efficient allocation,
there exists a policy such that
sincere picking with this policy
generates this allocation.
We can construct this policy using
the greedy algorithm in
the proof of Proposition 1 in 
\cite{bkrs2005}.
The reverse, however, is not true. 
Sincere picking may not return
a Pareto efficient allocation. 

\begin{remark}
Sincere picking can generate allocations
that are not Pareto efficient. 
\end{remark}
\myproof
Consider the policy
$1221$. 
Suppose the agents' utilitiea are as follows
$$
\begin{array}{c|ccccc}
& a& b & c & d \\ \hline
1& 5 &4 &2 &0 \\
2 & 8 & 2 & 1 & 0\\
\end{array}
$$
Both agents
have the same total utility
over the items. 
Sincere picking gives items $a$ and $d$ to agent 1
and items $b$ and $c$ to agent 2. 
This gives an utility of 5 to agent
1 and of 3 to agent 2. 
If they swap allocations, then
the utility of agent 1 increases to 6,
and of agent 2 to 8. Hence, 
sincere picking leads to
an allocation that is not Pareto efficient,
and does not have the optimal 
egalitarian or utilitarian social welfare. 
\myqed

We contrast this observation
with Proposition 1 in \cite{bkrs2005}.
This looks just at the rank of items
in an agent's preference ordering, ignoring
their precise utilities. 
Given two sets of items $S$ and $S'$ with $|S|=|S'|$,
an allocation of items $S$ to
an agent {\em dominates} the allocation
of items $S'$ iff for every item in $S-S'$
there is a different item in $S'-S$ that
is strictly less preferred.
They then define an
ordering, {\em ordinal efficiency} in terms of such domination. 
This is a strictly weaker ordering
than Pareto efficiency which is defined
in terms of utilities rather than ordinal rankings. 

Proposition 1 in \cite{bkrs2005}
demonstrates that 
ordinal efficiency corresponds {\em exactly}
to allocations generated by sequential 
allocation supposing sincere picking. 
On the other hand, only a subset of the allocations returned 
by sequential allocation are Pareto efficient.
And only a subset of these maximize the egalitarian social welfare.
We highlight the fact that there exists an allocation with the maximum possible egalitarian welfare.

\begin{remark}
There exists an allocation with the maximum possible
egalitarian
social welfare that is also Pareto efficient. 
\end{remark}
The argument is as follows. 
Among all allocations with maximum egalitarian welfare choose one with the largest utilitarian
welfare. This allocation is clearly Pareto efficient.

It follows quickly 
that there always exists a policy for sequential
allocation that gives an allocation
with the maximum possible egalitarian 
social welfare supposing sincere picking. 
Note that the proof does not rule out other
allocations which maximize egalitarian
social welfare which are not ordinal efficient,
and which cannot be generated by sequential allocation with 
sincere picking. 

\begin{myexample}
Suppose we have three agents ($1$ to $3$), three items ($a$ to $c$),
and Borda utilities. Let agent 1 have
a preference order $bac$, 
agent 2 have $abc$, and
agent 3 have $acb$. 
Then the allocation which gives
$a$ to agent 1, $b$ to agent 2 and
$c$ to agent 3 maximizes the egalitarian
social welfare. However, there is no
policy for sequential allocation that
will return such an allocation supposing
agents pick sincerely as no agent gets
a first choice item. 
\end{myexample}

Maximizing the utilitarian
social welfare also does not conflict with
Pareto efficiency. In this case, we point out the well-known fact that {\em any} allocation
that maximizes utilitarian social welfare is
Pareto efficient.

\begin{remark}
Any allocation with the maximum possible
utilitarian social welfare is also Pareto efficient. 
\end{remark}
The argument is as follows. 
Consider any allocation that has the maximum
possible utilitarian social welfare. 
Suppose there 
exists another allocation which Pareto improves it. 
Then the utility of every agent does not decrease.
This means that the sum of their utilities must
increase. This contradicts the assumption that
we have the maximum
possible utilitarian social welfare. 

Again it follows quickly that there exists a policy 
that gives an allocation
with the maximum possible utilitarian 
social welfare supposing sincere picking. 

\begin{figure}
\centering
\includegraphics[scale=0.4]{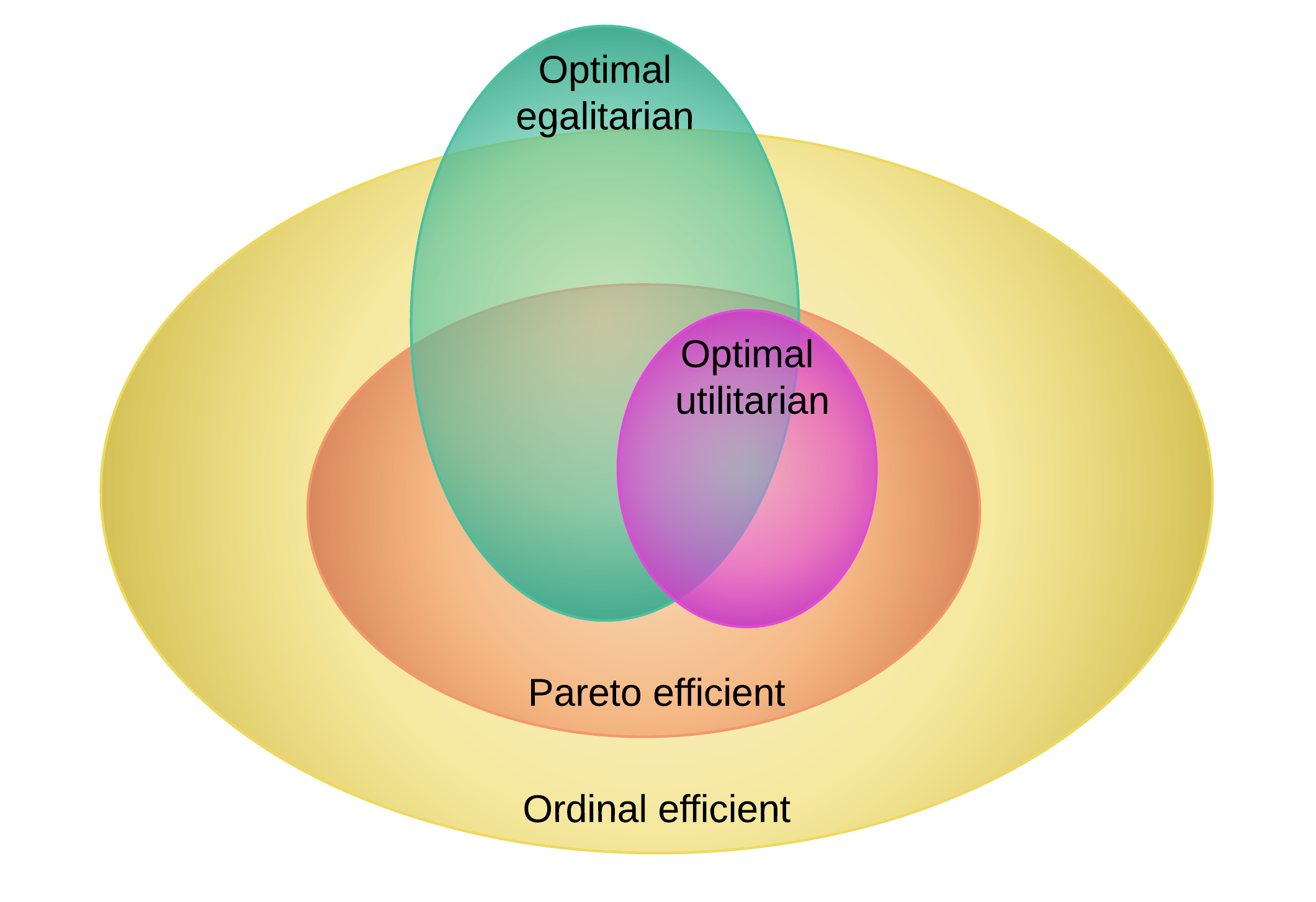}
\caption{Relationship between different allocations.
Sequential allocation with sincere picking can generate
any ordinal efficient allocation.}
\end{figure}

\section{Possible and necessary welfare}

Since sequential allocation may not
return allocations that are optimal from
either an egalitarian or utilitarian
perspective, we turn to the (computational) questions
of what social welfare is possible or necessary.
Note that throughout this paper, 
we suppose agents pick
sincerely. Whilst strategic
behaviour may be beneficial, risk averse agents
will tend to pick sincerely, especially
when the policy and/or utilities are private
information. Nevertheless, it would
be interesting future work to consider
agents acting strategically \cite{knwxaaai13}.

\vspace{0.2em}

\fbox{
\begin{minipage}{0.42\textwidth}
{\sc Possible/Necessary Utilitarian/ Egalitarian Welfare}

{\bf Input:} a set of $n$ items, $m$ agents each with 
utilities over the items, a class
of policies, and an integer $t$. 

{\bf Question:} Is there a policy/Does every policy
result in an allocation with 
an utilitarian/egalitarian social welfare of $t$ or greater
supposing agents pick items sincerely?
\end{minipage}
}

\vspace{0.2em}

The possible welfare questions answer a {\em policy control}
problem: can the chair choose a policy to achieve a 
given social welfare? Similar control problems have been
considered previously \cite{awxaaai15} but with the 
goal of allocating particular items to agents,
rather than, as here, of achieving a particular welfare. 
Note that we suppose we know
the (private) utilities of the agents. 
Our complexity results can be
seen as lower bounding the computational complexity
when we only have partial information about
the actual utilities. Alternatively, we may relax the
assumption that we know the actual utilities. 
For example,
as in \cite{peftd2003,indivisible,knwijcai13,bblnnrscomsoc2014}, 
we might suppose
that the utilities are simple functions of the ordinal
rank (e.g. Borda, lexicographical or quasi-indifferent
scores). As this is a special case of general utilities,
any result that control takes polynomial time in the general
case will map onto a polynomial time result in this more
restricted setting. 

When we prove that a particular possible or necessary
welfare problem takes polynomial time to solve, we will
typically do so by answering a closely related maximization
or minimization problem. Such problems are
interesting in their own right. 

\vspace{0.2em}

\fbox{
\begin{minipage}{0.42\textwidth}
{\sc Maximum/Minimum Utilitarian/ Egalitarian Welfare}

{\bf Input:} a set of $n$ items, $m$ agents each with 
utilities over the items, and a class
of policies. 

{\bf Output:} The maximum/minimum 
utilitarian/egalitarian social welfare possible
over all policies supposing agents pick items sincerely.
\end{minipage}
}

\vspace{0.2em}

\section{All possible policies}

If any policy is possible, it is easy
to maximize the utilitarian social welfare.
The chair just need 
to choose a policy that gives items to the agents which value
them most. 

\begin{mytheorem}
The {\sc Maximum} and {\sc Possible Utilitarian Welfare} 
problems are polynomial time solvable. 
\end{mytheorem}
\myproof
We order the items by the maximum utility assigned by
any agent. Ties can be broken in any way. We then
construct the policy that allocates items in this order
choosing the agent who gives an item the greater utility. 
No allocation can do better than this.
\myqed

The {\sc Minimum Egalitarian Welfare} problem also
takes polynomial time to solve. It is always zero. 
On the other hand, 
the {\sc Possible Egalitarian Welfare} problem is 
intractable in general, even in the special
case that all the agents have identical utilities for
the items. 

\begin{mytheorem} \label{epc}
The {\sc Possible Egalitarian Welfare} problem
for $m$ items and $n$ agents 
is strongly NP-complete when $m \geqslant 2n$.
\end{mytheorem}
\myproof
Membership in NP is shown by giving the policy. 
The proof uses a reduction from numerical 3-dimensional matching.  Given an integer $t$ and 3
multisets $X=\{x_1,\ldots,x_n\}$, $Y=\{y_1,\ldots,y_n\}$ and $Z=\{z_1,\ldots,z_n\}$ of integers with
$\sum_{i=1}^n (x_i+y_i+z_i)=nt$, this problem asks if there are permutations $\sigma$ and $\pi$ such
that $x_i+y_{\sigma(i)}+z_{\pi(i)} = t$ for all $i \in [n]$.  We construct an allocation problem
over $n$ agents and $m\geqslant 2n$ items as follows. Let $u=1+\sum_{i=1}^nz_i $. For every
$j\in[n]$, there is a ``big'' item with utility $u+x_i+y_j$ for agent $i$ ($i=1,\ldots,n$) and a
``small'' item which all agents give utility $z_j$. Finally, there are $m-2n$ items with zero utility for all
agents. We ask if we can achieve an egalitarian welfare of $u+t$. To achieve this, each agent 
must get precisely a utility of $u+t$. This is only possible if each agent gets one big item and one
small item, and $x_i+y_{\sigma(i)}+z_{\pi(i)}=t$ where $\sigma(i)$ and $\pi(i)$ denote are the
indices of the big and the small item obtained by agent $i$. Therefore, we can achieve the egalitarian welfare of $u+t$
iff there is a solution of the original numerical 3-dimensional matching problem.  \myqed

The same reduction proves that
the {\sc Maximum Egalitarian Welfare} problem
is NP-hard to compute. 
In the more restricted setting that utilities are 
Borda scores but agents have different ordinal
preferences, the 
{\sc Possible} and {\sc Maximum Egalitarian Welfare} problems
remain NP-hard. 
This follows from the reduction used to prove
Theorem 3 in \cite{bblnrseumas2013} (due 
to an anonymous reviewer of the paper).
It proves that deciding if there is an allocation 
with an egalitarian social welfare greater
than or equal to some constant $t$ is NP-complete even
when utilities are Borda scores. 
It is easy to show that there is a policy 
that finds the 
precise allocation
constructed in this reduction. 

\section{Balanced policies}

It might be considered unfair to 
use any policy, even one in which one agent
gets many more items than another. 
Whilst looking for allocations that maximize fairness and
efficiency, Brams and King \shortcite{bkrs2005} observe that 
{\em ``the symbolic value of giving players equal numbers
of items, such as landing slots at an airport, may be
important''}. 
We therefore consider the restricted
class of balanced policies. In a {\em balanced}
policy, each agent gets the same number of items.
For simplicity, we suppose the number of items
is an integer multiple of the number of agents
and add dummy items of no utility otherwise. 
Limiting sequential allocation to 
balanced policies impacts the social welfare
that can be obtained. 


To maximize utilitarian welfare, we cannot 
simply give items 
to the agents that value them most. This may
violate balance. 
Despite this restriction,
we can still find the policy that maximizes
the utilitarian welfare in polynomial time. 

\begin{mytheorem}
The {\sc Maximum} and {\sc Possible Utilitarian Welfare}
problems for balanced policies take polynomial time to solve. 
\end{mytheorem}
\myproof
We suppose that there are $kn$ items to
divide between the $n$ agents. 
We set up a min cost max flow problem.
We connect the source node to nodes representing
the agents, each with a capacity of $k$ and no cost. 
We connect the nodes representing agents to
nodes representing the items. Each edge has 
a capacity of 1, and a 
cost equal to minus the utility that the
agent assigns to the item. 
Finally we connect the nodes representing the items
to the target node, each with an edge of capacity 1 and
zero cost. 
We find a Pareto efficient allocation
from any such flow using the top trading cycle algorithm
\cite{ttc}. A policy can be constructed
that achieves this Pareto efficient
allocation by again exploiting Proposition 1 in
\cite{bkrs2005}. 
\myqed

By comparison,
the {\sc Necessary Utilitarian Welfare} problem is
intractable for balanced policies.

\begin{mytheorem} \label{nuw}
The {\sc Necessary Utilitarian Welfare} 
problem for balanced policies
is coNP-complete.
\end{mytheorem}
\myproof
We reduce from the {\sc Necessary Item} problem for
balanced policies which is coNP-complete
even when limited to an agent's most preferred
item \cite{awxaaai15}. 
Let one agent have
utility of 1 for their most preferred item, 
and zero utility for all others. By comparison,
let the other agents all have utility 
1 for every item. 
Then the {\sc Necessary Item} problem 
is equivalent to asking if an utilitarian
welfare of $m$ or more is necessary. 
\myqed

It follows that the {\sc Minimum Utilitarian Welfare} problem
for balanced policies is NP-hard to compute.
Restricting to balanced policies 
also does not change the 
NP-hardness of the {\sc Maximum} and {\sc Possible
Egalitarian Welfare} problems. 
This follows almost immediately from the
reduction used in the proof of Theorem \ref{epc}.
Note that this reduction uses 
policies in which some agents get less than 3 
items (which are not balanced). However, such 
unbalanced policies can be trivially ignored as they result
in poor egalitarian social welfare. Note also that
when a numerical 3-dimensional
matching exists, the corresponding successful policy 
constructed in the reduction is balanced. 
When utilities are specified in binary,
an easy reduction from the {\sc Equi-Partition}
problem demonstrates that the {\sc Possible Egalitarian Welfare}
problem restricted to balanced policies
is NP-complete even with just two agents who have
identical utilities. Finally, the {\sc Necessary Egalitarian
Welfare} problem is intractable for balanced policies.

\begin{mytheorem}
The {\sc Necessary Egalitarian Welfare} 
problem for balanced policies
is coNP-complete.
\end{mytheorem}
\myproof
The same reduction as in the proof of Theorem \ref{nuw}.
\myqed

\section{Recursively balanced policies}

Balanced policies might still be considered unfair. 
For example, a policy like $11112222$ favours the first
agent even though it is balanced,
and is guaranteed to return a Pareto efficient
allocation. We therefore consider
an even more restrictive class: recursively balanced
policies. In such a policy, 
items are allocated in rounds, and each agent appears once in
each round. For simplicity, we again suppose
that the number of items
is an integer multiple of the number of agents
and add dummy items of no utility otherwise. 
When the number of items equals the number of
agents, all balanced policies are recursively balanced. 
For this reason, we focus on problems where
the number of items exceeds the number of agents. 
Recursively balanced policies include 
the balanced alternating policy (12211221\ldots),
as well as the
Thue-Morse sequence (122121121221\ldots).
With two agents, recursively balanced policies
are concatenations of 12 and 21. 
Other simple properties of recursively
balanced policies follow immediately from
their definition. 
For example, no agent has more than two successive picks
in a recursively balanced policy.
Limiting sequential allocation to recursively 
balanced policies may further impact the social welfare
that can be obtained. 

There are several situations where focusing
on recursively balanced policies does not hurt welfare.
For example, with Borda utilities, 
the expected utilitarian social
welfare for two agents is not impacted by limiting allocation
to recursively balanced policies. The simple
alternating policy which is recursively balanced is
optimal in expectation \cite{knwijcai13}.
Similarly for Borda utilities and small $n$, 
the expected egalitarian social
welfare for two agents is not impacted. We have
computed the policies that maximize expected
egalitarian social welfare for up to 12 items and
for each $n$, at least one optimal policy
is recursively balanced. 

In general, restricting to recursively balanced
policies results in it being intractable
to decide if a given egalitarian or utilitarian
welfare can or must be achieved.  

\begin{mytheorem} \label{topk}
The {\sc Possible Egalitarian} and {\sc Possible Utilitarian Welfare} 
problems for recursively balanced policies
are NP-complete, 
whilst the 
{\sc Necessary Egalitarian} and {\sc Necessary Utilitarian Welfare} 
are coNP-complete. 
\end{mytheorem}
\myproof
We reduce from the corresponding
problem of deciding whether the top $k$
most preferred items of an agent are
possible or necessary \cite{awxaaai15}. 
The {\sc Top}-$k$ {\sc Possible Set} problem for
recursively balanced policies is
NP-complete for $k\geq 3$. 
We reduce this to the {\sc Possible Egalitarian Welfare}
problem as follows. Let one agent have
utility of $k^2$ for their $i$th most preferred items ($i\leq k$)
and zero utility for all others. By comparison,
let the other agents all have utility 
$k^3$ or greater for any item. 
Then the {\sc Top}-$k$ {\sc Possible Set} problem 
is equivalent to asking if an egalitarian
welfare of $k^3$ or more is possible. 
We also reduce 
the {\sc Top}-$k$ {\sc Possible Set} problem
to the {\sc Possible Utilitarian Welfare}
problem as follows. Let one agent have
utility of $mk^2$ for their $i$th most preferred items ($i\leq k$)
and zero utility for all others. By comparison,
let all the other agents have utility 
of $k$ or less for any item. 
Then the {\sc Top}-$k$ {\sc Possible Set} problem 
is equivalent to asking if an utilitarian
welfare of $mk^3$ or more is possible. 

The {\sc Top}-$k$ {\sc Necessary Set} problem is
coNP-complete for recursively balanced policies. 
We reduce this to the {\sc Necessary Egalitarian Welfare}
problem as follows. Let one agent have
total utility of $k^2$ for their $k$ most preferred items
and zero utility for all others. By comparison,
let the other agents all have utility 
$k^3$ or greater for any item. 
Then 
the {\sc Top}-$k$ {\sc Necessary Set} problem 
is equivalent to asking if an egalitarian welfare
of $k^2$ is necessary. 
We also reduce 
the {\sc Top}-$k$ {\sc Necessary Set} problem
to the {\sc Necessary Utilitarian Welfare}
problem as follows. Let one agent have
utility of $mk^2$ for their $i$th most preferred items ($i\leq k$)
and zero utility for all others. By comparison,
let all the other agents have utility 
of $k$ or less for any item. 
Then the {\sc Top}-$k$ {\sc Necessary Set} problem 
is equivalent to asking if an utilitarian
welfare of $mk^3$ or more is necessary.
\myqed

Even when agents have identical utilities, 
these problems can remain intractable. 

\begin{mytheorem}
When allocating $2n$ items between two agents, the 
\textsc{Possible Egalitarian Welfare} problem
for recursively balanced policies
is NP-complete even when agents have identical utilities given in binary.
\end{mytheorem}
\myproof Membership in NP is clear. For the hardness we use reduction from \textsc{Partition}: for
positive integers $a_1,\ldots,a_{n}$ with $a_1+\cdots+a_n=2B$, the problem is to decide if there is
a nonempty set $I\subseteq[n]$ with $\sum_{i\in I}a_i=B$. We reduce this to the \textsc{Possible
  Egalitarian Welfare} problem for two agents and $2n$ items with utilities $c_{1}=2B$, $c_{2n}=0$,
and
\[c_{2k}=c_{2k+1}=c_{2k-1}-a_k\qquad\text{for }k=1,2,\ldots,n-1.\] 
Let $C=\sum_{i=1}^{2n}c_i$ be the sum of the utilities. Note that an egalitarian welfare of $C/2$ is
equivalent to both agents achieving the same utility $u_1=u_2$. In round $k$, the items with
utilities $c_{2k-1}$ and $c_{2k}$ are allocated. From $c_{2k-1}-c_{2k}=a_k$ it follows that the
difference $u_1-u_2$ between the agents' utilities increases by $a_k$ if agent $1$ starts and
decreases if agent $2$ starts. Let $I\subseteq[n]$ be the set of rounds in which agent $1$
starts. An egalitarian social welfare of $C/2$ is achieved if and only if
\[0=u_1-u_2=\sum_{k\in I}a_k-\sum_{k\in[n]\setminus I}a_k,\]
i.e., if and only if there is a perfect {partition}. 
\myqed

\section{Balanced alternating policies}

The final and most restricted class of policies we
consider is that of balanced alternating. This is
the subclass of recursively balanced policies in
which each round is the reverse of the previous.
When allocating students to courses at the
Harvard Business School, such a policy is chosen
uniformly at random from the space of all possible
balanced alternating policies. This gives
a form of procedural fairness. 

\begin{mytheorem}
The {\sc Possible Egalitarian} and {\sc Possible Utilitarian Welfare} 
problems for balanced alternating policies
are NP-complete, 
whilst the 
{\sc Necessary Egalitarian} and {\sc Necessary Utilitarian Welfare} 
are coNP-complete. 
\end{mytheorem}
\myproof
By reduction as in the proof of Theorem \ref{topk}
from the corresponding {\sc Top}-$k$ {\sc Possible} or
{\sc Necessary Set} problem restricted to balanced alternating
policies. The {\sc Top}-$k$ {\sc Possible} problem
for balanced alternating
policies is NP-complete for $k\geq 2$, whilst 
the {\sc Top}-$k$ {\sc Necessary Set} problem is coNP-complete
\cite{awxaaai15}. 
\myqed

It follows that it is NP-hard to compute the probability 
that the Harvard Business School course allocation
mechanism returns an allocation with egalitarian or utilitarian
welfare greater than or equal to some given value, $t$. 

\section{Two agents}

We now consider some special cases which
are more tractable. With two agents, we can find a balanced
policy that maximizes the egalitarian 
or utilitarian welfare
in polynomial time.

\begin{mytheorem}
The {\sc Maximum Egalitarian}
and {\sc Maximum Utilitarian Welfare} problems
with balanced policies 
can be be solved in $O(k^2n^3)$ and $O(kn^2)$ time
respectively 
when allocating $2n$ items between two agents
with utilities (that may be different)
taken from $[0,k]$. 
\end{mytheorem}
\myproof
We put the items into some (arbitrary) order
and consider how each item is allocated in turn. 
We construct a $2n$ step dynamic program
in which the $i$th step corresponds to
the decision of where to allocate
the $i$th item in this order.
The states of this dynamic program
are triples containing
the number of items
allocated to the first agent, 
the sum of the utilities of the items so far
allocated to the first agent, and
the sum of the utilities of the items so far
allocated to the second agent. We can compute
the number of items allocated
to the second agent from this. 
As both  sums are bounded in size by $2kn$, this
dynamic program has $O(k^2n^3)$ states.
For the maximum utilitarian welfare,
the states of the dynamic program can be simpler
and just need to be pairs containing
the number of items
allocated to the first agent, and
the sum of the utilities of the items so far
allocated to both agents.
\myqed

This result generalizes to
a bounded number of agents. 
On the other hand, when utilities are specified in
binary, an easy reduction from the {\sc Partition}
problem demonstrates that the {\sc Possible Egalitarian Welfare}
problem is NP-complete even when the two agents have
identical utilities. 
This is almost identical to Proposition 2 in \cite{indivisible}
which shows that deciding if there is 
a policy that ensures a given {expected} egalitarian
welfare is NP-complete when the utilities
of the two agents are identical.

\begin{table*}[htb]
{
\begin{center}
\begin{tabular}{|r|r|r|r|r|} \hline
 & all policies & balanced & recursively balanced & balanced alternating \\ \hline
{\sc Possible Egalitarian Welfare} & NPC &  NPC & NPC & NPC \\ \hline
{\sc Possible Utilitarian Welfare} & P & P & NPC & NPC \\  \hline
{\sc Necessary Egalitarian Welfare} & P & coNPC & coNPC & coNPC \\  \hline
{\sc Necessary Utilitarian Welfare} & ? & coNPC & coNPC & coNPC \\  \hline
\end{tabular}
\end{center}
}
\caption{Summary of results: NPC=NP-complete, P=polynomial.}
\end{table*}

With recursively balanced policies, 
we consider the case
where agents have the same ordinal ranking over items.

\begin{mytheorem}
The {\sc Maximum} and {\sc Possible Egalitarian Welfare}
problems for recursively balanced policies
can be solved in $O(k^2n^2)$, whilst the 
{\sc Maximum} and {\sc Possible Utilitarian Welfare} problems
can be solved in just $O(kn)$ time 
when allocating $2n$ items between two agents
when agents have the same ordering over items
but possibly different utilities, and
utilities are drawn from $[0,k]$. 
\end{mytheorem}
\myproof
We construct a $n$ step dynamic program
in which each step corresponds to
one round of allocating one item to each
of the agents. The states of this dynamic program
are pairs containing the sums of the utilities of items so far
allocated to the two agents. As both 
sums are bounded by $kn$, this
dynamic program has $O(k^2n^2)$ states.
To compute the optimal utilitarian social
welfare, we can use a simpler dynamic program
where the states are just
the sum of the utilities 
allocated to the two agents.
\myqed

This result again generalizes
to a bounded number of agents easily. 

\section{House allocation}

Another more tractable case is house
allocation, when we have only as many items as agents. 
In this case, we can solve the {\sc Maximum} and 
{\sc Possible Egalitarian Welfare}
problems over all possible
policies in polynomial time. We construct a graph between
agents and items with edges for all items that have a utility 
greater than or equal to the desired egalitarian social 
welfare. The {\sc Possible Egalitarian Welfare} problem is solvable
if we can find a perfect matching in this graph. To construct
a satisfying policy, we find a Pareto efficient allocation
from this matching using the top trading cycle algorithm
\cite{ttc}.
A policy can be constructed
that achieves this Pareto efficient
allocation using Proposition 1 in
\cite{bkrs2005}.
This tractability of this case suggests an interesting
open problem. 
We have proved that {\sc Possible Egalitarian Welfare}
problem is NP-complete for $m=2n$ but takes polynomial time for
$m=n$. 
This leaves open the complexity in between. 

\section{Other related work}

As mentioned earlier,
Bouveret and Lang \shortcite{indivisible} consider the case in which the
utilities of items are simply functions of the ordinal rankings.
They prove that
any recursively balanced policy 
tends to an allocation giving
the optimal {expected} egalitarian or utilitarian social welfare
as the number of items grows,
supposing sincere picking, utilities
that are Borda scores and all ordinal rankings
being equiprobable. 
In addition, they compute the optimal
policies for maximizing the 
expected egalitarian or utilitarian social welfare
under the same assumptions for up to 12 items.
The optimal policies for two agents and an
even number of items are recursively balanced.
Kalinowski, Narodytska and Walsh  \shortcite{knwijcai13} 
prove that the alternating policy maximizes the
expected utilitarian social welfare under these 
same assumptions. We again note
that such results are about maximizing the {\em expected} welfare
supposing limited knowledge about the
utilities, whilst the results here about maximizing
the {\em exact} welfare supposing the chair
knows the {\em actual} utilities. 

There has been some study of strategic behaviour of agents
(as opposed to the chair) in the sequential allocation mechanism.
It can, for example, be viewed
as a repeated game. When all agents have complete
information, we can compute the subgame perfect Nash
equilibrium. This is unique and takes polynomial time to compute
for two agents \cite{kcor71,knwxaaai13}, 
but for an arbitrary
number of agents, there can be an exponential
number of equilibria and computing even one
is PSPACE-hard \cite{knwxaaai13}.
More recently, Bouveret and Lang  \shortcite{blecai14}
consider how an agent or coalition of 
agents can strategically mis-report their
preferences in a sequential allocation mechanism
supposing the other agents
act sincerely. 
They show that the loss of social
welfare caused by such manipulation is
not great. For example, with Borda scoring, 
two agents, and the alternating policy, there
was at most a 33\% loss in the utilitarian welfare.

More recently, a family of rules for dividing
indivisible goods among agents has been proposed
that take as input the agents' ordinal rankings
over the items, a scoring vector, and a social
welfare aggregation function \cite{bblnrseumas2013,bblnnrscomsoc2014}. 
They return the allocation that maximizes the
social welfare according to this scoring rule and 
aggregation function. Whilst such rules have a number of desirable 
properties like monotonicity, they have a number
of less desirable properties including a high
computational complexity to compute the actual
allocation (unless we have a bounded number of agents
in which case we can typically use dynamic programming).
This contrasts with sequential allocation where
computing the allocation take just linear time. 
Baumeister {\it et al.} 
\shortcite{bblnnrscomsoc2014} also compute the 
multiplicative/additive ``price of elicitation-freeness'',
the worst-case ratio/difference in social welfare between such
allocations and the allocation returned by sequential allocation.
Whilst their results are limited
to simple alternating policies, the prices are typically
not great. For example, the optimal utilitarian social
welfare with Borda scores
is at most twice that returned by sequential
allocation using simple alternation. 

\section{Conclusions}

We have considered the implications on social
welfare of choosing different policies when
using a sequential mechanism to 
allocate indivisible goods. In particular,
we consider the (computational) questions
of what welfare is possible or necessary. 
The former is related to the control problem
in which a (benevolent) chair chooses a policy for the
sequential allocation mechanism to improve
the social welfare. 
These questions are also related to the
expected welfare when we choose a policy
uniformly at random. 
Our results are summarized in Table 1.
There are many interesting open questions.
For example, how difficult is it to find
a recursively balanced policy that 
returns a Pareto efficient allocation supposing
agents pick sincerely?


\bibliographystyle{aaai}


\end{document}